\documentclass[conference]{IEEEtran}
\ifCLASSINFOpdf
\else
\fi
\usepackage{amsthm}
\newtheorem{theorem}{Theorem} 

\usepackage{amsmath}
\usepackage{multirow}
\usepackage{bbold}
\usepackage{graphicx}
\usepackage{array}
\usepackage{mdwmath}
\usepackage{mdwtab}
\usepackage{eqparbox}
\newcolumntype{P}[1]{>{\centering\arraybackslash}p{#1}}

\begin{document}
%
\title{Subclass Knowledge Distillation with Known Subclass Labels}


%
\author{\IEEEauthorblockN{Ahmad Sajedi\IEEEauthorrefmark{1},
Yuri A. Lawryshyn\IEEEauthorrefmark{2} and
Konstantinos N. Plataniotis\IEEEauthorrefmark{1}}
\IEEEauthorblockA{\IEEEauthorrefmark{1} The Edward S. Rogers Sr. Department of Electrical \& Computer Engineering, University of Toronto\\}
\IEEEauthorblockA{\IEEEauthorrefmark{2} Centre for Management of Technology \& Entrepreneurship (CMTE), University of Toronto\\
Emails: ahmad.sajedi@mail.utoronto.ca, yuri.lawryshyn@utoronto.ca, kostas@ece.utoronto.ca}\\}


\maketitle
\begin{abstract}
\boldmath
This work introduces a novel knowledge distillation framework for classification tasks where information on existing subclasses is available and taken into consideration. In classification tasks with a small number of classes or binary detection, the amount of information transferred from the teacher to the student is restricted, thus limiting the utility of knowledge distillation. Performance can be improved by leveraging information of possible subclasses within the classes. To that end, we propose the so-called Subclass Knowledge Distillation (SKD), a process of transferring the knowledge of predicted subclasses from a teacher to a smaller student. Meaningful information that is not in the teacher's class logits but exists in subclass logits (e.g., similarities within classes) will be conveyed to the student through the SKD, which will then boost the student's performance. Analytically, we measure how much extra information the teacher can provide the student via the SKD to demonstrate the efficacy of our work. The framework developed is evaluated in clinical application, namely colorectal polyp binary classification. It is a practical problem with two classes and a number of subclasses per class. In this application, clinician-provided annotations are used to define subclasses based on the annotation label's variability in a curriculum style of learning. A lightweight, low-complexity student trained with the SKD framework achieves an F1-score of $85.05\%$, an improvement of $1.47\%$, and a $2.10\%$ gain over the student that is trained with and without conventional knowledge distillation, respectively. The $2.10\%$ F1-score gap between students trained with and without the SKD can be explained by the extra subclass knowledge, i.e., the extra $0.4656$ label bits per sample that the teacher can transfer in our experiment. The SKD framework can benefit from using more information to increase student performance, but it comes at the expense of the availability of subclass labels.
\end{abstract}



%
\IEEEpeerreviewmaketitle

\section{Introduction}\label{sec1}
In many real-world classification problems, each class has a number of available semantically meaningful subclasses. For example, in the cancer diagnosis task, which involves the detection of benign and abnormal lesions, the abnormal class may have multiple subclasses in which each of them can express different types or organs of cancer disease \cite{oakden2020hidden, mlynarski2019deep}. Models trained exclusively on class labels often ignore the fine-grained knowledge of subclasses, which can have an effect on model training, particularly for clinical tasks such as cancer detection \cite{oakden2020hidden, sohoni2020no}. We can take advantage of this subclass knowledge by forcing the model to learn subclass labels. Then, the knowledge from the classes and subclasses can be transferred from one model (teacher) to another (student) using the Knowledge Distillation (KD) framework for the purpose of model compression \cite{kd}.
\begin{figure} 
    \centering
    \includegraphics[width= 1\linewidth]{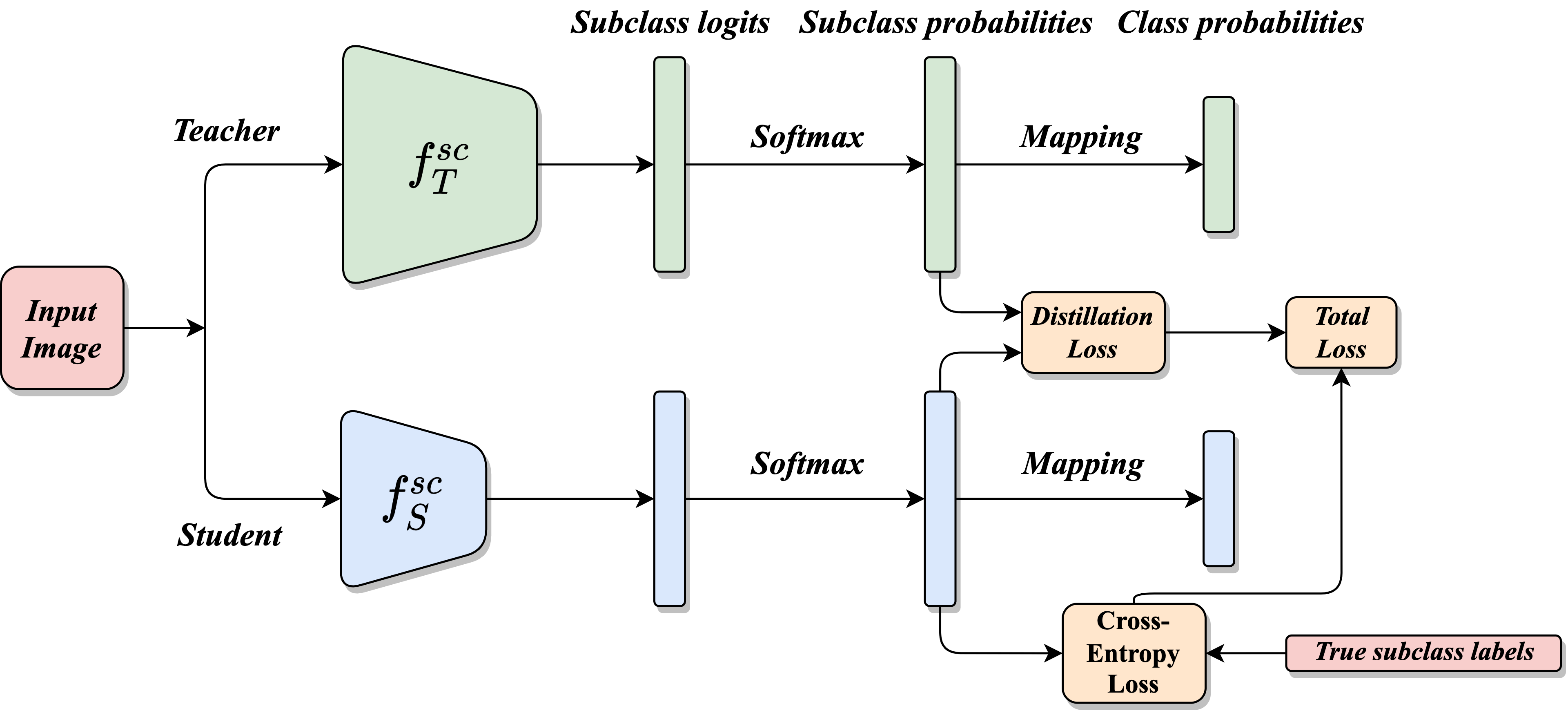}
    \caption{Subclass Knowledge Distillation framework.}
    \label{fig1}
    \vspace{-0.45cm}
\end{figure}

The relative probabilities of incorrect class prediction (i.e., dark knowledge) can reveal a lot about the teacher's generalization tendencies. Dark knowledge can be extracted from the probability distribution of soft targets and used during conventional KD \cite{kd, fitnet}. As long as we distill the teacher's knowledge using soft logits at a high temperature, the amount of information the teacher generalizes is linear in the number of classes \cite{skd}. When datasets contain many classes, knowledge transfer from teacher to student is typically successful, as the teacher has more relevant information about the function being taught \cite{skd}. Meanwhile, in classification tasks with a few classes or binary detection problems, the amount of information available to the student is restricted, thus limiting the utility of the KD. To address this problem, we can leverage hidden subclass knowledge, the knowledge of available subclasses that is not captured in the teacher's class logits.

Müller {\em et al.} \cite{skd} compelled the teacher model to create artificial subclasses for each class during the training phase with auxiliary contrastive loss. The student model is then trained to mimic the invented teacher's subclass predictions (probabilities). This paper discusses "model-induced" subclasses, which can also result in non-meaningful subclasses. Tzelepi {\em et al.} \cite{oskd, oskd2} proposed Online Subclass Knowledge Distillation (OSKD) to estimate a set of subgroups and then train the lightweight model using a self-distillation approach. These subgroups are estimated based on the different numbers of nearest neighbours in each sample to reveal the similarities inside classes. They showed that revealing estimated subclass knowledge improves KD, but they provided no analytical justification for their findings. Furthermore, the assumption that the nearest neighbours of each sample inside a class share the same semantic meaningful similarities is not always true, especially in high-dimensional spaces \cite{bengio2005curse}. Unlike previous methods of subclass distillation in the literature, we propose the Subclass Knowledge Distillation (SKD) framework to transfer the knowledge of known and available subclasses within each class. Here, we are talking about meaningful "problem-induced" subclasses that already exist. Our research also aims to bridge the gap between analytical and semantic explanation in a knowledge distillation framework.
The following is a summary of our contributions:
\vspace{-0.1cm}
\begin{itemize}
    \item Propose the SKD, a novel framework to efficiently distill subclass knowledge into the student network and further boost its performance.
    \vspace{-0.1cm}
    \item Analyze how much information the student can learn about the teacher's generalization through the SKD framework.
    \vspace{-0.1cm}
    \item Conduct an experimental study on the MHIST dataset, a clinically important binary dataset, to evaluate the performance of the SKD framework. Our experimental results demonstrate that the learned subclass factorization increases student performance.
    \vspace{-0.1cm}
\end{itemize}
\section{Subclass Knowledge Distillation} \label{sec2}
Subclass Knowledge Distillation is expected to make use of subclass knowledge when performing classification tasks with a small number of classes. In the following, we will elaborate on the details of the SKD framework in a teacher-student context where subclasses are known and available. 
\vspace{-.14cm}
\subsection{SKD Framework} \label{subsec2.1}
Knowledge distillation is used to abstract the representation obtained in a high-complexity model into a simpler model to maintain good performance but in a more concise architecture. In classification problems, the abstraction of the representation may be compromised if only a small number of classes are available \cite{skd}. In this case, one can exploit the additional information from postulated subclasses within the original classes for a robust transfer.
Given a teacher model $T$ and a student model $S$, let us have the one-hot vector of subclass label $y_{i}^{sc}$ corresponding to the training sample $x_{i}$ of sample space $\mathcal{X}$. Suppose the mapping from subclass to class labels is also known (Fig. \ref{fig1}). In the SKD framework, unlike Müller's work \cite{skd}, the teacher is trained using the ground-truth supervision of \textbf{subclass labels} by minimizing the following cross-entropy loss (CE) associated with subclass probabilities:
\vspace{-.1cm}
\begin{align} \label{eq1}
 \mathcal{L}_{teacher} = \sum_{x_{i}\in \mathcal{X}}CE(\sigma(f_{T}^{sc}(x_{i})), y_{i}^{sc}),
\end{align}
where the teacher $T$ computes subclass logits and probabilities using the function $f_{T}^{sc}(.)$ and softmax functions $\sigma(.)$, respectively. The SKD framework is a process in which the student is trained to mimic the teacher’s behavior. However, instead of leveraging the original class labels, the student learns to match the teacher's subclass prediction by optimizing the following SKD loss:
\vspace{-.1cm}
\begin{flalign} \label{eq2}
  \mathcal{L}_{skd} = \sum_{x_{i}\in \mathcal{X}} KL(\sigma(\frac{f_{T}^{sc}(x_{i})}{\tau}), \sigma(\frac{f_{S}^{sc}(x_{i})}{\tau})),
\end{flalign}
where KL stands for Kullback-Leibler divergence and $f_S^{sc}(.)$ denotes the function used to compute subclass logits in the student model $S$. The temperature hyperparameter, $\tau$ is used to generate soft subclass predictions while controlling the entropy of the output distribution. We use a linear combination of the SKD loss, $\mathcal{L}_{skd}$, and the standard cross-entropy loss, $\mathcal{L}_{ce}$ as the objective function for training the student model:
\begin{flalign} \label{eq3}
 \mathcal{L}_{student} = \lambda \mathcal{L}_{ce} + (1-\lambda) \mathcal{L}_{skd},
\end{flalign}
where $\mathcal{L}_{ce} = \sum_{x_{i}\in \mathcal{X}}CE(\sigma(f_{S}^{sc}(x_{i})), y_{i}^{sc})$ is the cross-entropy loss and $\lambda \in [0, 1]$ is a task balance hyper-parameter. Following the supervision of subclass labels to train the teacher and student, class output probabilities can be determined simply by adding the probabilities of all subclasses within the class. It is worth noting that, while the teacher and student were trained on subclass labels, they are evaluated on class labels (Fig. \ref{fig1}). In contrast to other works \cite{skd, oskd, oskd2}, the postulated subclasses in the SKD framework are semantically meaningful and related to the tasks, i.e., they are not artificially created during training but are the result of expert annotations. The SKD framework is useful because meaningful subclass labels impose finer supervision for feature learning than class labels. These fine-grained labels can help the teacher learn more features and generalize better when evaluating class labels.
\vspace{-.1cm}
\subsection{Analytical Measurement on the SKD Framework} \label{subse2.2}
To demonstrate the effectiveness of our framework, we calculate the number of label bits that the teacher can provide to the student using different types of discrete memoryless channels. The information theory channel is a system whose output is probabilistically dependent on its input \cite{cover, el2011network}. Every channel is defined by an input alphabet, an output alphabet, and a description of how the output depends on the input. In this paper, the true label space and the predicted label space are the input and output alphabets of our channel, namely $\mathcal{A}$ and $\mathcal{\hat{A}}$, respectively. Similar to the channel transition matrix in information theory, the normalized confusion matrix on the training set illustrates the relationship between the predicted and true labels of the teacher network. Furthermore, and most importantly, the information capacity of each channel indicates the amount of information it transmits, which is equivalent to the information label bits that the teacher can provide to the student in our study. It should be mentioned that all channels used in the paper are memoryless, as each predicted label is influenced only by the corresponding true label and not by earlier true or predicted labels.
\begin{figure} 
    \centering
    \includegraphics[width= 0.95\linewidth]{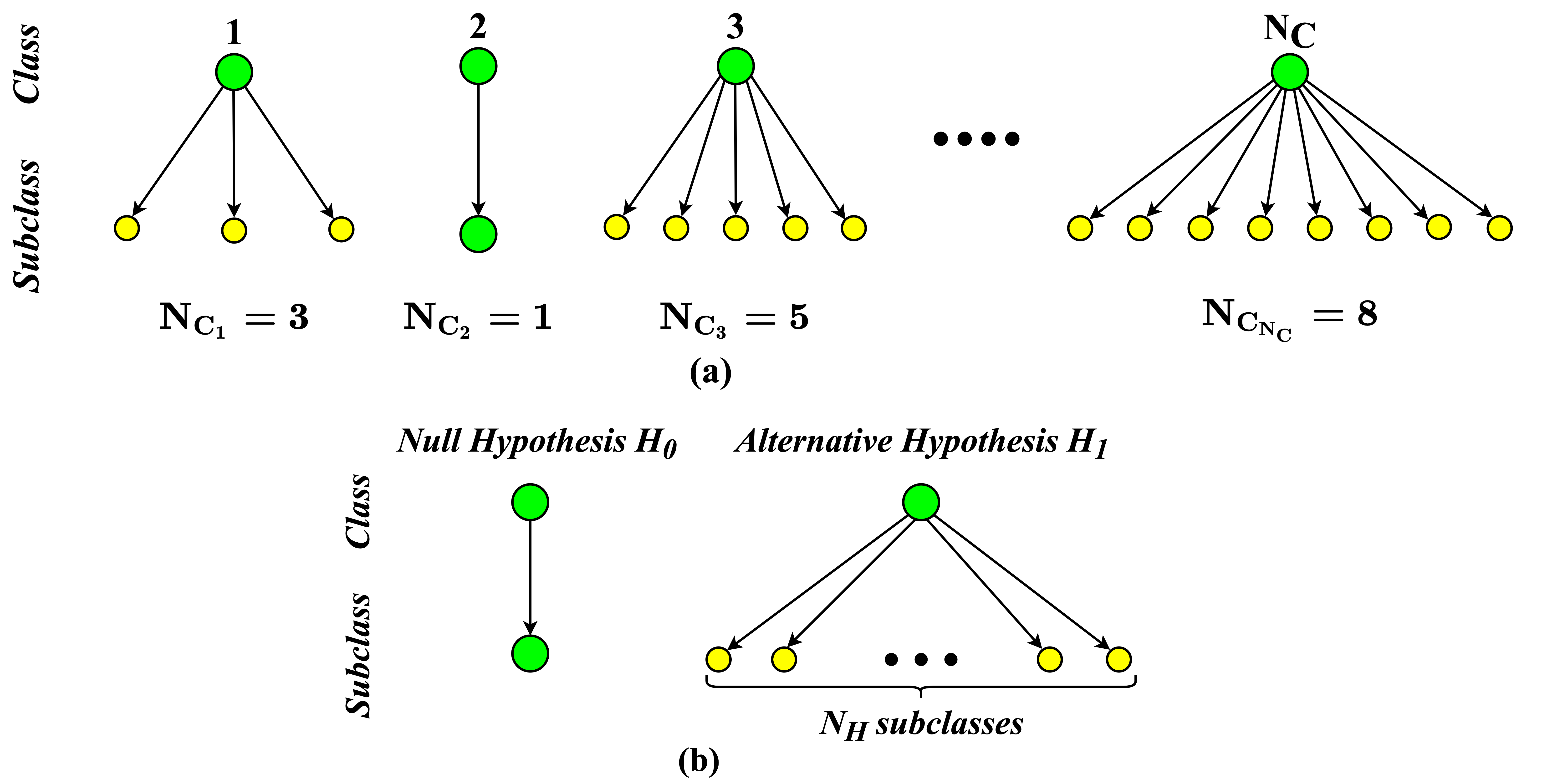}
    \caption{(a) Example of a class hierarchy in a multiclass classification task; (b) Class hierarchy of our binary detection problem.}
    \label{fig2}
    \vspace{-0.4cm}
\end{figure}
In the following theorem, we measured how many label bits per sample the teacher can transfer through the SKD when each class may contain a different number of subclasses and each subclass has multiple training samples (Fig. \ref{fig2} (a)).
\begin{theorem} \label{th1}
Suppose that the dataset has $N_{C}$ classes, and class $i$ contains known and available $N_{C_{i}}$ subclasses for $i$ in $\{1, 2, ..., N_{C}\}$. Let the teacher network predict each class $i$ (each subclass of class $i$, resp.) correctly with a probability of $P_{C}$ ($P_{C_{i}}$, resp.) during the training phase, while the remaining errors are distributed equally over the remaining $(N_{C}-1)$ classes ($(N_{C_{i}}-1)$ subclasses, resp.)  (Fig. \ref{fig3} (c)). Then, the number of label information bits the teacher can transfer through SKD is bounded above by
\vspace{-0.17cm}
\begin{flalign} \label{eq7}
 [\log N_{C} + P_{C}\log P_{C} +& (1-P_{C}) \log  \frac{1-P_{C}}{N_{C}-1}] + \nonumber\\
   [\sum_{k = 1}^{N_{C}} \frac{\sum_{j = 1}^{N_{C_{k}}} N_{S_{kj}}}{\sum_{i = 1}^{N_{C}} \sum_{j = 1}^{N_{C_{i}}} N_{S_{ij}}}& (log N_{C_{k}} + P_{C_{k}} log P_{C_{k}} + \nonumber\\ &(1-P_{C_{k}}) log \frac{1-P_{C_{k}}}{N_{C_{k}}-1}],
\end{flalign}
where the first and second brackets correspond to the class and subclass labels, respectively. The variable $N_{S_{ij}}$ indicates the number of training samples for subclass $j$ of class $i$. All information quantities are represented in bits, and the $\log$ function is to base $2$.
\end{theorem}
\begin{figure} 
    \centering
    \includegraphics[width= 0.921\linewidth]{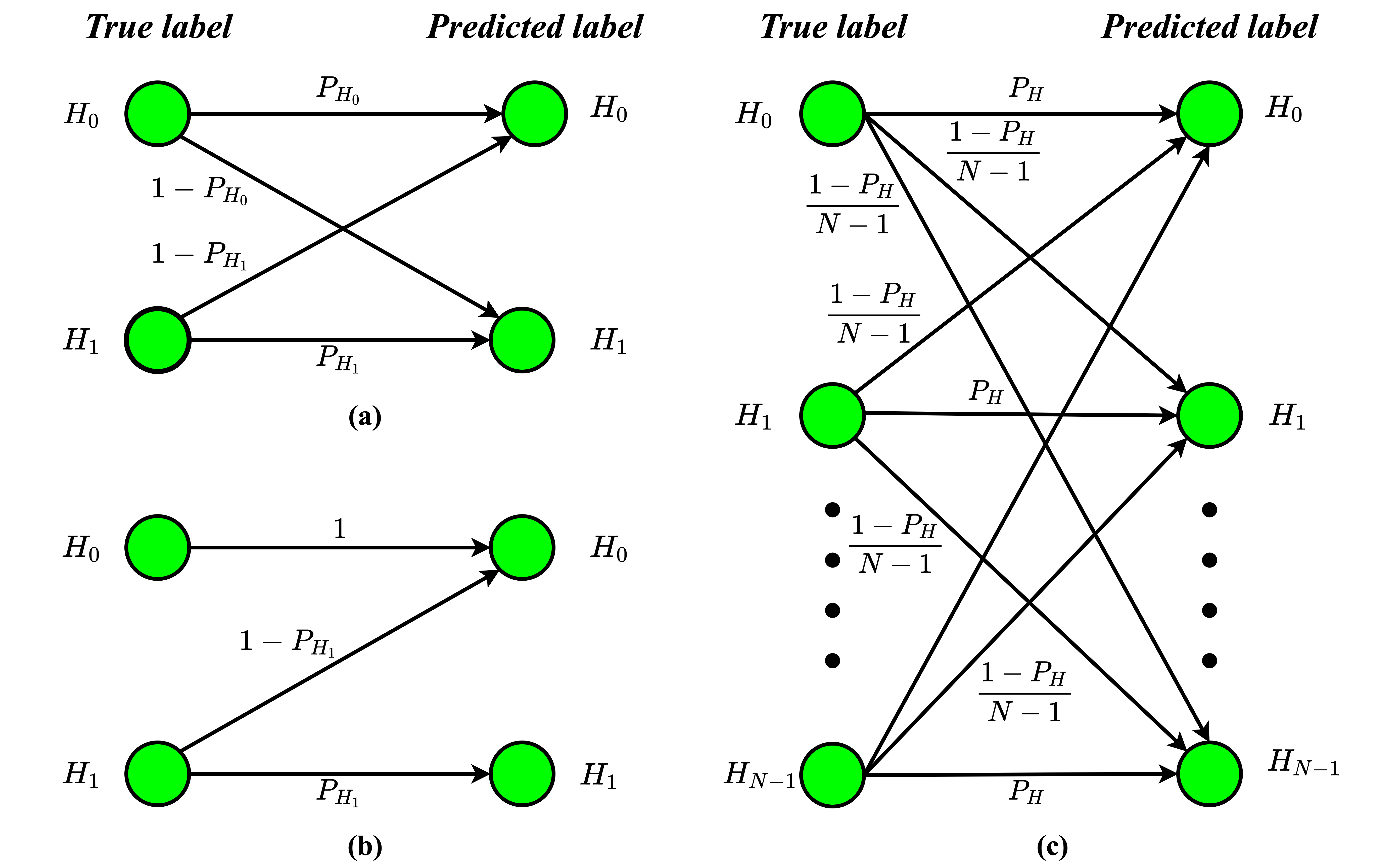}
    \caption{Discrete memoryless channels as potential models for quantifying the label bits that teacher can provide for the student: (a) Binary Asymmetric Channel; (b) Z-Channel; (c) Q-ary Symmetric Channel, where $N$ indicates the cardinality of input alphabet.}
    \label{fig3}
    \vspace{-0.4cm}
\end{figure}
\vspace{-0.2cm}
Within the problem of data-driven classification, there is a case where it is extremely important, which is the so-called detection case. The detection task is primarily a binary classification of the hypothesis under consideration, with the outcome typically being either a null hypothesis $H_{0}$ or an alternative hypothesis $H_{1}$. The detection problem is practically important because, for example, this is the case when someone tries to detect whether a person has cancer or not. In cancer diagnosis tasks, alternative hypothesis $H_{1}$ (abnormal class) may have $N_{H}$ subclasses in which each of them can express different types or organs of cancer disease \cite{oakden2020hidden, mlynarski2019deep, hosseini2019atlas, Dua:2019, bhattacharjee2001classification, esteva2017dermatologist} (Fig. \ref{fig2} (b)). Furthermore, it is fair to say that the majority of training samples are identified as normal class \cite{imbalance}, resulting in a biased dataset. For this binary detection task, the following theorem establishes an upper bound on the number of label bits that the teacher can transfer through the SKD. 
\begin{theorem} \label{th2}
Let the teacher network predict the null and alternative hypothesis correctly with a probability of $P_{H_{0}}$ and $P_{H_{1}}$, respectively (Fig. \ref{fig3} (a)). Suppose that the teacher predicts each subclass label of the alternative hypothesis properly with a probability of $P_{S}$ and the remaining errors are equally distributed throughout the remaining $(N_{S}-1)$ subclasses (Fig. \ref{fig3} (c)). Then the average number of label bits per training sample that the teacher can provide is bounded above by
\begin{flalign} \nonumber 
& [\log (1+2^{K(P_{H_{0}}, P_{H_{1}})}) -P_{H_{0}}K(P_{H_{0}}, P_{H_{1}}) - H_{b}(P_{H_{0}})] + \nonumber\\
 & [\frac{N_{H_{1}}}{N_{H_{0}}+N_{H_{1}}} (\log  N_{S}+P_{S}\log P_{S} +  (1-P_{S})\log  \frac{1-P_{S}}{N_{S}-1})],\nonumber 
\end{flalign}
where $N_{H_{0}}$ and $N_{H_{1}}$ represent the number of training samples in the null and alternative hypotheses, respectively. Note that $H_{b}(x) = -x\log{x} - (1-x)\log(1-x)$ is a binary entropy function and $K(P_{H_{0}}, P_{H_{1}}) = \frac{H_{b}(P_{H_{1}}) - H_{b}(P_{H_{0}})}{P_{H_{0}}+P_{H_{1}}-1}$.
\vspace{-0.25cm}
\end{theorem}
It is important to note that in the SKD framework, unlike previous works \cite{skd, oskd}, not only the amount of information carried by the labels matters but also the relevance of the labels to the task, which is determined by the expert annotators.
\vspace{-.25cm}
\section{Experiments and Results} \label{sec3}
\vspace{-.14cm}
In this section, we describe the experimental setups that will be used throughout the study. Our objective is to compress a large-scale teacher with high accuracy into a smaller student that is more appropriate for deployment on the MHIST dataset. For this aim, we rely on the distillation of subclass knowledge.
\vspace{-.68cm}
\subsection{Minimalist HIStopathology (MHIST) Dataset} \label{subsec3.1}
\vspace{-.1cm}
In this work, we focus on the clinically important classification problem between Hyperplastic Polyps (HPs) and Sessile Serrated Adenomas (SSAs) \cite{ssa, ssa_lesion, hp} on the MHIST dataset \cite{mhist}. HPs are generally benign, but SSAs are precancerous lesions that might progress to malignancy and require more frequent follow-up exams \cite{colonpolyps}. Pathologically, HPs have superficial serrations in the upper parts of the crypt, whereas in SSAs, serrations extend deeper into the crypt and the crypts are broad-based and may have a boot shape \cite{ssa2}. In the annotation phase of the MHIST dataset, seven practicing board-certified gastrointestinal pathologists separately and independently classified each of the $3,152$ images as either HP or SSA \cite{mhist}. The gold standard label was then allocated to each image on the basis of the majority vote among the seven labels, a common choice in literature \cite{breastmr, colorectaldeep}. 
In the MHIST, each class can be partitioned into $4$ subgroups according to the discrete level of difficulty, which is determined by image-level annotator agreement: (I) very easy to predict ($7/7$ annotator agreement), (II) easy to predict ($6/7$ annotator agreement), (III) hard to predict ($5/7$ annotator agreement), and (IV) very hard to predict ($4/7$ annotator agreement). These clinician-provided annotations are used to define subclasses based on the annotation label's variability in a curriculum style of learning \cite{wei2021learn}. Then, we take the following classification tasks:\vspace{0.14cm}\\  
$\mathbf{1.}$  \textbf{SubclassLevel-}$\mathbf{21}$\textbf{:} SSA has $2$ subclasses: (I) very easy and easy to predict, (II) hard and very hard to predict. HP has one subclass, which is the class itself.  \vspace{0.12cm}\\ 
$\mathbf{2.}$ \textbf{SubclassLevel-}$\mathbf{22}$\textbf{:} SSA and HP have $2$ subclasses: (I) very easy and easy to predict, (II) hard and very hard to predict.\vspace{0.12cm}\\ 
$\mathbf{3.}$ \textbf{SubclassLevel-}$\mathbf{12}$\textbf{:} SSA has a single subclass, which is the class itself. HP has $2$ subclasses: (I) very easy and easy to predict, (II) hard and very hard to predict.  

\begin{table*}[htb] 
\centering
\caption{Hyper-parameter details. (SL: SubclassLevel)}
\begin{tabular}{c c c c c c c c c}
\hline
{Task} & {Model} & {Optimizer} & {\# of epochs} & {Batch size} & {lr}  & {Weight decay} & {Temperature $\tau$} & {Task balance $\lambda$}\\
\hline
\multirow{2}{*}{SL-$21$} & {Teacher (ResNet50V2 \cite{resnet})} & Adam \cite{adam} & $80$ & $32$ & $0.0001$ & $0.0005$ & | & | \\
& {Student (NASNetMobile \cite{nasnetmobile})} & Adam  & $30$ & $32$ & $0.001$ & $0.0005$ & $5$ (SKD) -- $128$ (KD) &  $0.45$ (SKD) -- $0.45$ (KD)\\
\hline
\multirow{2}{*}{SL-$22$} & {Teacher (ResNet50V2)} & Adam & $80$ & $32$ & $0.0001$ & $0.0005$ & | & | \\
& {Student (NASNetMobile)} & Adam & $30$ & $32$ & $0.001$ & $0.0005$ & 5 (SKD) -- 128 (KD) &  0.75 (SKD) -- 0.45 (KD)\\
\hline
\multirow{2}{*}{SL-$12$} & {Teacher (ResNet50V2)} & Adam & $80$ & $32$ & $0.0001$ & $0.0005$ & | & | \\
& {Student (NASNetMobile)} & Adam & $30$ & $32$ & $0.0001$ & $0.0005$ & 5 (SKD) -- 128 (KD) &  0.45 (SKD) -- 0.45 (KD) \\
\hline
\end{tabular}
\label{table1}
\vspace{-0.42cm}
\end{table*}
\vspace{-0.25cm}
\subsection{Evaluation Metrics} \label{subsec3.2}
\vspace{-0.15cm}
In this section, we use the MHIST dataset to run experiments on the principles described in the previous sections. The MHIST dataset is skewed in favour of the HP ($2162$ samples for HP, $990$ samples for SSA). When a dataset is unbalanced, it is critical to strike a balance between precision and recall. As a result, we use the F1-score to compare the performance of the models in our experiment \cite{wang2021transpath}. Because our models were trained using random initialization, we ran each model $60$ times and reported the mean and standard deviation of its F1-score as an evaluation metric. In all experiments, teacher and student models were trained on subclass or class labels but evaluated only on class labels.
\vspace{-.32cm}
\subsection{Experimental Setups} \label{subsec3.3}
\vspace{-.12cm}
We trained the ResNet50V2 network \cite{resnet} and the NASNetMobile network \cite{nasnetmobile} to be used as teacher and student models, respectively. We used $5$-fold cross-validation to tune the hyper-parameters, where the hyper-parameter details are reported in Table \ref{table1}. For the teacher and student, we used data augmentation, a learning rate decay factor of $0.91$, a dropout with a probability of $0.2$ in the final softmax matrix, and all other parameters were left as default \cite{tensorflow2015-whitepaper}. All of the models in this paper were trained on the NVIDIA Tesla V100-SXM2-16GB \cite{teslav100} using the TensorFlow V2 framework.

\vspace{-.32cm}
\subsection{Experimental Results} \label{subsec3.4}
\vspace{-.12cm}
We report the overall test performances in Table \ref{table2}. We started by training the teacher network on the MHIST dataset for all three tasks when class and/or subclass labels were available. As a baseline, we trained the student network without knowledge distillation when only class labels were known. Then we investigate how the distillation of knowledge can help the student perform better. In particular, in the SL-$12$ task, the student trained with conventional KD received a $0.63\%$ increase in F1-score in comparison to the baseline student. Following that, we distilled subclass knowledge from a teacher that had been trained with subclass labels and found a class F1-score improvement when compared to the student with conventional KD and from scratch. In the SL-$12$ task, the student model that had been trained with the SKD framework achieved an F1-score of $85.05\%$, an improvement of $1.47\%$ and $2.10\%$ over the students that were trained with conventional KD and from scratch, respectively. The same results for the other tasks can be observed in Table \ref{table2}. Thus, in our clinical setup, the SKD can compress a large-scale teacher into a smaller, less computationally complex student without severely sacrificing performance (the best teacher using subclass labels improved the F1-score by only $0.92 \%$ when compared to the best student using the SKD). To be more precise, we measured the computational cost of teacher and student models trained with SKD in the SL-$12$ task using the number of multiply-adds (FLOPs) \cite{xie2017aggregated} and the number of trainable parameters. As shown in Table \ref{table3}, the computational complexity of the student network is $6$x less than that of the teacher, while its inference time is roughly equal to the teacher's inference time.


\begin{table}[htb] 
\centering
\caption{Results of the test F1-score in different tasks on MHIST. The baseline corresponds to training the student on class labels without distillation. The standard deviation is calculated over 60 runs. (MSD: Model-induced Subclass Distillation \cite{skd}, SL: SubclassLevel)}
\begin{tabular}{P{1.2cm} P{3.8cm} c}
\hline
{Task} & {Method} & {Binary Class F1-score(\%)}\\


\hline

\multirow{5}{*}{SL-$21$} & {Teacher (using class labels)} & $85.53\pm0.84$\\
& {Teacher (using subclass labels)} & $85.78\pm0.99$\\
& {Student (baseline) (using class labels)} & $\mathbf{82.87\pm0.94}$\\
& {Student (using subclass labels)} & $83.47\pm1.84$\\
& {Student + KD} & $83.53\pm1.57$\\
& {Student + SKD} & $\mathbf{84.52\pm1.54}$\\
\hline


\multirow{6}{*}{SL-$22$}& {Teacher (using class labels)} & $85.49\pm0.78$\\
& {Teacher (using subclass labels)} & $85.75\pm0.94$\\
& {Student (baseline) (using class labels)} & $\mathbf{82.92\pm1.01}$\\
& {Student (using subclass labels)} & $83.89\pm1.48$\\
& {Student + KD} & $83.49\pm1.73$\\
& {Student + MSD} & ${84.03\pm1.65}$\\
& {Student + SKD} & $\mathbf{84.94\pm1.34}$\\
\hline

\multirow{5}{*}{SL-$12$}& {Teacher (using class labels)} & $85.60\pm0.78$\\
& {Teacher (using subclass labels)} & $85.97\pm0.87$\\ 
& {Student (baseline) (using class labels)} & $\mathbf{82.95\pm1.01}$\\
& {Student (using subclass labels)} & $84.16\pm1.75$\\
& {Student + KD} & $83.58\pm1.62$\\
& {Student + SKD} & $\mathbf{85.05\pm1.48}$\\
\hline

\end{tabular}
\label{table2}
\vspace{-.1cm}
\end{table}

\begin{table}[htb]
\centering
\caption{Results of computational cost (G-FLOPs), interference time, and the trainable parameters for the teacher and the student networks trained in the SL-$12$ task.}
\begin{tabular}{P{1cm} P{1cm} P{2cm} P{1.5cm}}
\hline
{Model} & {FLOPs} & {Inference time} & Parameters\\
\hline
{Teacher} & $6.970$G & $5.29$ms & $20.57$M\\
{Student} & $\mathbf{1.136}$\textbf{G} &  $6.81$ms & $\mathbf{2.21}$ \textbf{M}\\
\hline
\end{tabular}
\label{table3}
\vspace{-.55cm}
\end{table}
\vspace{-.1cm}
Finally, we measured the label bits that the teacher can transfer to the student to show how the SKD framework benefits from subclass knowledge to improve the student's performance in colorectal polyps classification. The results in Table \ref{table4} show that the student, trained on the SKD, can gain $0.4656$ extra label bits per sample from hidden subclass knowledge. The difference in the number of label bits explains the $2.10\%$ F1-score gap between the students trained with and without the SKD in the binary classification task. Note that comparing the total label bits that a teacher can provide for a student lets us detect what level of sub-classification would be beneficial (e.g., SL-$12$ in our experiment).
\vspace{-.12cm}
\begin{table}[htb]
\centering
\caption{The upper bound on the number of label bits per sample that the teacher can provide in different tasks. Total label bits is the summation of class and subclass label bits per sample. (In the ClassLevel task, only class labels are known.)}
\begin{tabular}{c c c c}
\hline
{Task} & Class label bits & Subclass label bits & Total label bits\\
\hline
{ClassLevel} & $0.8363$ & | & $0.8363$ \\
{SL-$21$} & $0.7915$ & $0.3749$ & $1.1664$  \\
{SL-$22$} & $0.5781$ & $0.6977$ & $1.2758$ \\
{SL-$12$} & $0.8793$ & $0.4226$ & $\mathbf{1.3019}$ \\
\hline
\end{tabular}
\label{table4}
\vspace{-.1cm}
\end{table}
\vspace{-.50cm}
\section{Conclusion and Future Works} \label{sec4}
\vspace{-.10cm}
Subclass Knowledge Distillation is proposed in this paper for classification tasks where information on existing subclasses is available and taken into consideration. We showed that we can improve the performance of the lightweight student by transferring hidden subclass knowledge, the additional meaningful information that helps the teacher learn more fine-grained features. This extra knowledge is then analytically measured using channel capacity concepts from the field of information theory. Finally, the SKD was evaluated in the clinical binary classification and showed that it can benefit from subclass knowledge to boost student performance. Future work could be theoretically, such as investigating the proposed upper bound's tightness, or experimentally, like evaluating SKD on more datasets, even if the subclasses are unknown.

\bibliographystyle{IEEEtran}

\bibliography{refs}
%

%
%

\onecolumn
\def\x{{\mathbf x}}
\def\L{{\cal L}}

\begin{center}
\textbf{\LARGE Supplementary Materials}
\end{center}

\section{Proof of Theorem 1}

Based on the assumptions of Therem $1$, the normalized class label confusion matrix of the training set follows the structural pattern of the Q-ary Symmetric Channel's transition matrix. Thus, the information capacity of the Q-ary Symmetric Channel indicates the maximum number of label bits that a teacher can convey to a student via its class labels. This capability is achieved through a uniform distribution across the class label space, as specified by
\begin{flalign} \label{eq3-8}
\log N_{C} - H(P_{C}, \frac{1-P_{C}}{N_{C}-1}, \ldots, \frac{1-P_{C}}{N_{C}-1}) = \log N_{C} + P_{C} \log P_{C} + (1-P_{C}) \log \frac{1-P_{C}}{N_{C}-1} 
\end{flalign}
Unless the relative frequencies of training samples across classes match the capacity-achieving distributions, the Q-ary Symmetric Channel's capacity will be an upper bound on the number of label bits that the teacher can provide using class labels. In other words, if the relative frequencies of training samples over class labels converge to a uniform distribution, the upper bound will be tight and will approach the real label bits.\\
In parallel with the class labels, the Q-ary Symmetric Channel could also be a suitable model to analyze the subclass label bits for a given class $i$, as the normalized subclass label confusion matrix for the training set within class $i$ follows the structural pattern of the Q-ary Symmetric Channel's transition matrix. Given that each subclass has a different number of training samples, the following weighted average can be used to further establish an upper bound on the number of subclass label bits per sample that the teacher can provide.  
\begin{flalign} \label{eq3-9}
&\frac{\sum_{j = 1}^{N_{C_{1}}} N_{S_{1j}}}{\sum_{i = 1}^{N_{C}} \sum_{j = 1}^{N_{C_{i}}} N_{S_{ij}}} (\log N_{C_{1}} - H(P_{C_{1}}, \frac{1-P_{C_{1}}}{N_{C_{1}}-1}, \frac{1-P_{C_{1}}}{N_{C_{1}}-1}, \ldots, \frac{1-P_{C_{1}}}{N_{C_{1}}-1})+ \nonumber\\
&\frac{\sum_{j = 1}^{N_{C_{2}}} N_{S_{2j}}}{\sum_{i = 1}^{N_{C}} \sum_{j = 1}^{N_{C_{i}}} N_{S_{ij}}} (\log N_{C_{2}} - H(P_{C_{2}}, \frac{1-P_{C_{2}}}{N_{C_{2}}-1}, \frac{1-P_{C_{2}}}{N_{C_{2}}-1}, \ldots, \frac{1-P_{C_{2}}}{N_{C_{2}}-1}))+ \ldots + \nonumber\\
&\frac{\sum_{j = 1}^{N_{C_{N_{C}}}} N_{S_{N_{C}j}}}{\sum_{i = 1}^{N_{C}} \sum_{j = 1}^{N_{C_{i}}} N_{S_{ij}}} (\log N_{C_{N_{C}}} - H(P_{C_{N_{C}}}, \frac{1-P_{C_{N_{C}}}}{N_{C_{N_{C}}}-1}, \frac{1-P_{C_{N_{C}}}}{N_{C_{N_{C}}}-1}, \ldots, \frac{1-P_{C_{N_{C}}}}{N_{C_{N_{C}}}-1})) = \nonumber \\
&\sum_{k = 1}^{N_{C}} \frac{\sum_{j = 1}^{N_{C_{k}}} N_{S_{kj}}}{\sum_{i = 1}^{N_{C}} \sum_{j = 1}^{N_{C_{i}}} N_{S_{ij}}} (\log N_{C_{k}} - H(P_{C_{k}}, \frac{1-P_{C_{k}}}{N_{C_{k}}-1}, \frac{1-P_{C_{k}}}{N_{C_{k}}-1}, \ldots, \frac{1-P_{C_{2}}}{N_{C_{2}}-1})) = \nonumber \\
& \sum_{k = 1}^{N_{C}} \frac{\sum_{j = 1}^{N_{C_{k}}} N_{S_{kj}}}{\sum_{i = 1}^{N_{C}} \sum_{j = 1}^{N_{C_{i}}} N_{S_{ij}}} (\log N_{C_{k}} + P_{C_{k}} \log P_{C_{k}} + (1-P_{C_{k}}) \log \frac{1-P_{C_{k}}}{N_{C_{k}}-1})
\end{flalign}
where $\frac{\sum_{j = 1}^{N_{C_{i}}} N_{S_{ij}}}{\sum_{i = 1}^{N_{C}} \sum_{j = 1}^{N_{C_{i}}} N_{S_{ij}}}$ is the weight applied to the subclass label bits of class $i$. Finally, the summation of class and subclass label information bits, i.e., Equations \ref{eq3-8} and \ref{eq3-9} completes the proof of Theorem $1$.

\section{Proof of Theorem 2}
Given that the task is binary detection and the teacher predicts both classes with different probabilities in general, the normalized confusion matrix for the class classification task will follow the structural pattern of the channel transition matrix for Binary Asymmetric Channel (BAC) (Fig. 3(a) of the paper). As a result, the information capacity of BAC gives us the maximum number of label bits that the teacher can convey to the student via class labels. Let $Y$ and $\hat{Y}$ be random variables taking values in $\mathcal{A}$ and $\mathcal{\hat{A}}$, respectively. Without loss of generality, we can assume that $P_{H_{1}} \leq P_{H_{0}}$ and $\alpha$ is the probability of event that the training sample belongs to the alternative hypothesis. The capacity of BAC is then given by      
\begin{flalign} \label{eq3-11}
C_{BAC} & = \max_{\alpha}   I(Y;\hat{Y}) \overset{(a)}= \max_{\alpha}  (H(\hat{Y})-H(\hat{Y}|Y))   \nonumber \\
  &\overset{(b)}= \max_{\alpha} [H_{b}(\alpha P_{H_{1}} + (1-\alpha)(1-P_{H_{0}})) -  (1-\alpha)H_{b}(P_{H_{0}}) - \alpha H_{b}(P_{H_{1}})] 
\end{flalign}
where $H(\hat{Y}|Y)$ denotes the conditional entropy of $\hat{Y}$ given $Y$; $(a)$ and $(b)$ are followed by the definition of mutual information $I(Y;\hat{Y})$ and the mutual information corresponding to BAC, respectively. To determine the optimal point, we calculate the derivative of the cost function with respect to $\alpha$ and, after some simplifications, obtain
\begin{flalign} \label{eq3-12}
 \alpha^{*} = \frac{1}{(P_{H_{0}}+P_{H_{1}}-1)} [\frac{1}{2^{K(P_{H_{0}}, P_{H_{1}})}+1}-(1-P_{H_{0}})]
\end{flalign}
where $K(P_{H_{0}}, P_{H_{1}}) = \frac{H_{b}(P_{H_{1}})-H_{b}(P_{H_{0}})}{P_{H_{0}}+P_{H_{1}}-1}$. The capacity of BAC is then calculated by substituting $\alpha = \alpha^{*}$ into the cost function of Equation \ref{eq3-11}.
\begin{flalign}  \label{eq3-13}
C_{BAC} = \log (1+2^{K(P_{H_{0}}, P_{H_{1}})})-P_{H_{0}}K(P_{H_{0}}, P_{H_{1}}) - H_{b}(P_{H_{0}}) 
\end{flalign}
Unless the relative frequencies of training samples over classes match the capacity-achieving distributions, the capacity of BAC will be an upper bound on the number of label bits provided by the teacher's class labels. In other words, when the relative frequency of training samples for alternative hypothesis approaches $\alpha^{*}$, the upper bound becomes tighter and converges to the real label bits.\\
Parallel to proof of Theorem $1$, Q-ary Symmetric Channel could be a suitable model to analyze the subclass label bits because the normalized subclass label confusion matrix follows the structural pattern of Q-ary Symmetric Channel's transition matrix (Fig. 3(c) of the paper: $N=N_{H}$ and $P_{H} = P_{H_{11}}$). Q-ary symmetric channel capacity, in conjunction with the proof of Theorem $1$, gives us the desired upper bound on the number of subclass label bits provided by the teacher.
\begin{flalign} \label{eq3-14}
[\frac{N_{H_{0}}}{N_{H_{0}}+N_{H_{1}}} & \times 0] + [\frac{N_{H_{1}}}{N_{H_{0}}+N_{H_{1}}} \times (\log N_{S}+{P_{S}}\log P_{S} + (1-P_{S})\log \frac{1-P_{S}}{N_{S}-1})] = \nonumber \\ 
& \frac{N_{H_{1}}}{N_{H_{0}}+N_{H_{1}}} \times (\log N_{S}+{P_{S}}\log P_{S} + (1-P_{S})\log \frac{1-P_{S}}{N_{S}-1})
\end{flalign}
where $\frac{N_{H_{1}}}{N_{H_{0}}+N_{H_{1}}}$ ($\frac{N_{H_{0}}}{N_{H_{0}}+N_{H_{1}}}$, resp.) denotes the relative frequency of training samples for alternative hypothesis (null hypothesis, resp.). It is worth noting that the null hypothesis contains only one subclass, which is itself. Finally, the summation of upper bounds on class and subclass label bits completes the proof of Theorem $2$.

\end{document}